\PassOptionsToPackage{sort}{natbib}
 
 \documentclass[pmlr,twocolumn]{jmlr} 

\usepackage{booktabs}
\usepackage[load-configurations=version-1]{siunitx} 
 \usepackage{hyperref}
 
 \usepackage{microtype}
 \usepackage{paralist}

\theorembodyfont{\upshape}
\theoremheaderfont{\scshape}
\theorempostheader{:}
\theoremsep{\newline}

\jmlrvolume{ML4H Extended Abstract arXiv Index}
\jmlryear{2020}
\jmlrsubmitted{2020}
\jmlrpublished{}
\jmlrworkshop{Machine Learning for Health (ML4H) 2020} 

\title[Image analysis for Alzheimer's disease prediction]{Image analysis for Alzheimer's disease prediction: Embracing pathological hallmarks for model architecture design}

 \author{%
\Name{Sarah C. Br\"uningk}$^\ast$, \Name{Felix Hensel}$^\ast$, \Name{Catherine R. Jutzeler**}, 
\Name{Bastian Rieck**}\\
\addr{\footnotesize Department of Biosystems Science and Engineering, ETH Zurich and SIB Swiss Institute of Bioinformatics.}  \hfill \emph{\footnotesize{$\ast$:~These authors contributed equally. **:~These authors share last authorship.}
}
}

\begin{document}

\maketitle

\begin{abstract}
Alzheimer's disease (AD) is associated with local~(e.g.~brain tissue atrophy) and global brain changes~(loss of cerebral connectivity), which can be detected by high-resolution structural magnetic resonance imaging. Conventionally, these changes and their relation to AD are investigated independently. Here, we introduce a novel, highly-scalable approach that simultaneously captures \emph{local} and \emph{global} changes in the diseased brain. It is based on a neural network architecture that combines patch-based, high-resolution 3D--CNNs with global topological features, evaluating multi-scale brain tissue connectivity. Our local--global approach reached competitive results with an average precision score of $0.95\pm0.03$ for the classification of cognitively normal subjects and AD patients~(prevalence $\approx55\%$).
\end{abstract}

\section{Introduction}
\label{sec:intro}
Affecting an estimated 35 million people worldwide, Alzheimer’s disease (AD) is the leading cause ($\approx$70\%) of dementia in elderly people \citep{winblad2016defeating}. AD is a chronic progressive disease, neuropathologically hallmarked by amyloid plaques, neurofibrillary tangles, glial responses, as well as neuronal and synaptic loss \citep{serrano2011neuropathological, scheff2006hippocampal,suemoto2017neuropathological}. As the disease progresses, these pathological processes trigger large-scale changes to the brain morphology, including atrophy, loss of cerebral connectivity, and volumetric shrinkage of distinct brain areas (e.g.~amygdala and hippocampus). Using magnetic resonance imaging (MRI), these morphological changes can be quantified \citep{rosenbloom2008magnetic}, both locally at high resolution (small scale atrophy) and globally throughout the image (tissue connectivity).
Machine learning has frequently been applied for MRI image classification and discovery of early imaging biomarkers of AD. Convolutional Neural Networks (CNNs) are widely used for these tasks \citep{Jo2019}. Owing to the high resolution three dimensional (3D) MRI images, training of deep 3D--CNNs on the full image space is computationally expensive. Alternatively, MRI images have been down-sampled at the cost of lower resolution \citep{Oh2019,Korolev2017,Jin2019} or patch-based CNNs have been applied \citep{Ahmed2019, Lin2018, Liu2018, Liu2018b} that are limited in detecting global, whole-image features.
We propose to address the trade-off between image resolution and computational cost through topological data analysis (TDA). TDA is a rapidly developing field based on algebraic topology that aims to study the shape of complex data. One of its flagship tools is \emph{persistent homology} (PH).
Enjoying stability properties and robustness against noise, PH captures multi-scale information on the connectivity of a data set, such as connected components, cycles, and higher-dimensional voids. In recent years, TDA has been successfully combined with machine learning methods; e.g.~\citet{Hofer-DLwTS}.
The aim of this study is to combine local and global neuroimaging features to capture the full spectrum of AD-induced brain changes with the overarching aim of improving the understanding of this disease and to provide a scalable solution to the challenge of 3D--image analysis. Two approaches, an ensemble of patch-based 3D--CNNs, and a combination of TDA with a single patch 3D--CNN were investigated. A deliberately simple CNN architecture was chosen as baseline in this preliminary study.

\section{Methods}

\paragraph{Data selection and preprocessing}

We included structural T1-weighted, MR images (no contrast) from the \href{http://adni.loni.usc.edu/}{Alzheimer's Disease Neuroimaging Initiative} (ADNI) for AD patients and healthy controls (CN) of matched age groups. Data from all ADNI subcohorts (ADNI1, 2, 3, and GO) were used. For details on the data selection and preprocessing, see \appendixref{apd:preprocessing}.

\begin{figure*}[tbp]
\floatconts
  {fig:mixed-CNN}
  {\caption{a) TDA 2D- and 3D-CNN architectures with ensemble model ii) using the preclassification layer encodings in a fully connected layer.
 b) APS (five run average, top) and normalised, centred class probabilities for one AD patient (single run, bottom) serving as features in ensemble ii) overlaid to a sagittal slice of the MR image. The image patch giving the best classification accuracy is highlighted.}}
  {\includegraphics[width=0.9\textwidth]{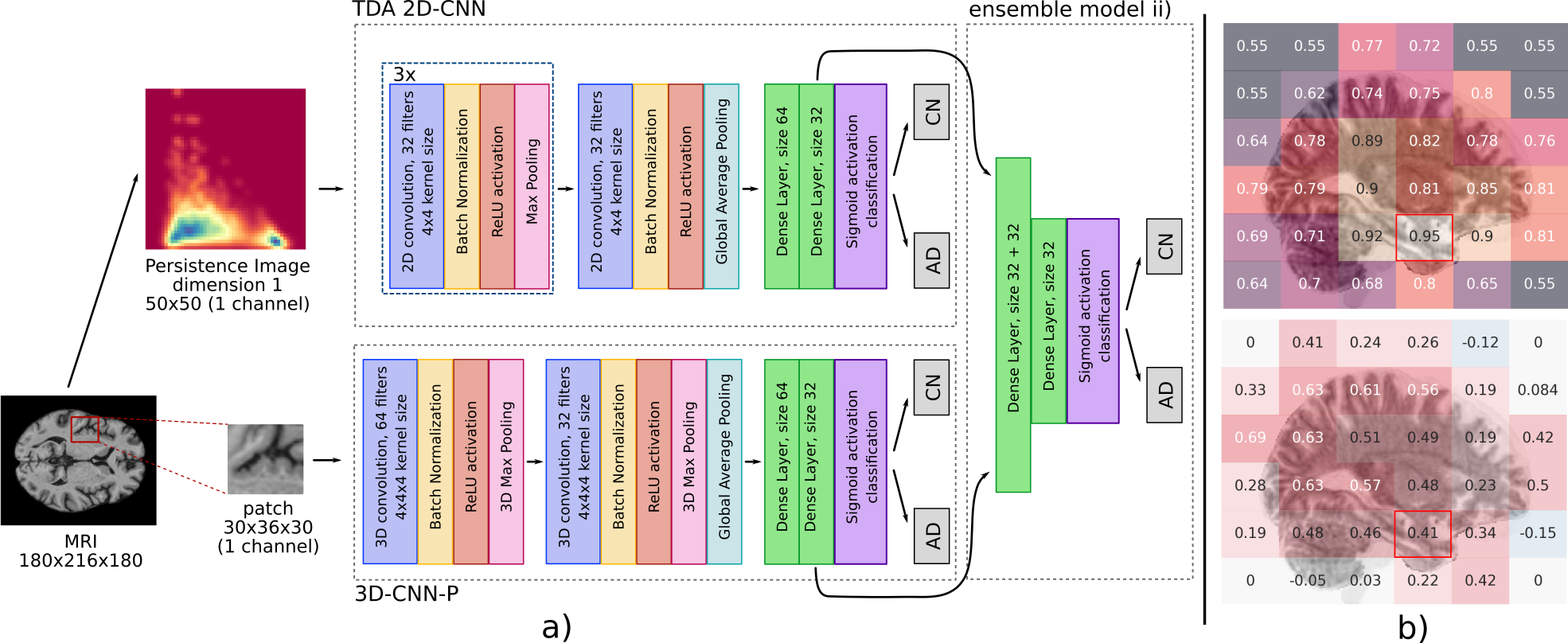}}
\end{figure*}

\paragraph{Convolutional neural network}
Preprocessed images were rescaled ($\mathbin{{193}{\times}{229}{\times}{193}}$ voxels), cropped ($\mathbin{{180}{\times}{216}{\times}{180}}$), and cut into $216$ patches of size $\mathbin{{30}{\times}{36}{\times}{30}}$ - a trade-off between image detail captured and computational cost. Hyperparameters of a 3D-CNN (see \figureref{fig:mixed-CNN}, 3DCNN--P) were optimised by random sampling on a patch within the hippocampus, denoted as 3DCNN--P*. Models using these optimal hyperparameters were trained independently on each image patch within 2500 epochs with early stopping based on validation loss using the Adam optimiser with He uniform initialisation. Our method is implemented in Python using Keras/TensorFlow; we plan on making our implementation publicly available.

\paragraph{Topological features}
For 3D MR images, topological features can occur in dimensions~0~(connected components), 1~(cycles), and 2~(voids).
We follow recent work~\citep{Rieck20}, which proposed:
\begin{inparaenum}
\item calculating persistent homology, and
\item vectorising the resulting descriptors via \emph{persistence images}~\citep[PI]{Adams-PI},
\end{inparaenum}
thus simplifying their integration into neural networks; we refer to \citet{Edelsbrunner-Harer-10} for a comprehensive introduction to TDA. Classification based on topological features was performed using a 2D-CNN as outlined in \autoref{fig:mixed-CNN}. Following hyperparameter optimisation, only \mbox{1-dimensional} features were used, as they yielded the best performance. We tested several preprocessing steps on the MRI images~(Gaussian filtering, down-sampling, filtering out low-persistence features, calculating both sublevel-sets and superlevel-sets) and of these report the best results to date.

\paragraph{Ensemble models}
Two approaches were considered that combined local and global imaging features through using i) all 216 image patches, or ii) a single image patch with PI representation of the full image. For i), class probabilities for each subject and model were normalised, and centred around a decision boundary of zero and served as input for logistic regression (LR) with grid search optimised hyperparameters.
For ii), the preclassification encodings of the individually trained 2D/3D--CNNs (3D--CNN--P* and 2D--CNN-PI) were combined through a single fully connected layer with optimized l1 regularization and sigmoidal activation (see \autoref{fig:mixed-CNN}). 

\paragraph{Performance evaluation}
We performed $4$-fold cross validation~(CV) splitting patients into training~($80\%$) and validation~($20\%$) sets, stratified by class labels. All longitudinal images per subject were used for training and a single image, randomly selected from the longitudinal set, was selected for each validation patient. Depending on the split, 851--895/291-298 train and 73--74/73--74 validation images/subjects were used with an average validation prevalence of AD of $54.6\pm0.5\%$.
Independent of the model architecture, training was repeated four times (runs) for each CV fold. We averaged all metrics over individual runs, then over folds, reporting mean values with standard deviations. A Wilcoxon signed rank test was performed to evaluate statistical significance.

\section{Results}
\tableref{tab:results} gives an overview of the classification performance of the different architectures investigated. The performance of a single patch based 3D--CNN was strongly influenced by the anatomical location of the image patch as shown in \figureref{fig:mixed-CNN}b). The best-performing model (denoted as 3D--CNN--P*) was obtained for a patch at the base of the brain encompassing parts of the left hippocampus and amygdala, two structures whose changes have previously been reported as potential AD imaging biomarkers \citep{Poulin2011}. Patches containing parts of other AD associated brain structures, such as the thalamus, caudate, or the ventricles also yielded competitive classification results.
Classification performance of any TDA-based model alone was influenced by the model PI preprocessing. So far, the best results in terms of APS were achieved with a superlevel PI input ($\text{APS} = 0.79\pm0.04$). None of the tested TDA-based approaches was superior to 3D-CNN-P*.
Ensemble model i) combining the predictions of all 216 3D--CNN--Ps had an improved classification performance in terms of APS relative to the best single patch model, 3D--CNN--P*, yet differences were not significant ($p > 0.05$). So far, no statistically significant difference between 3D--CNN--P* and the 3D--CNN--P*--TDA ensemble ii) were observed ($p > 0.05$).

\begin{table*}[tbp]
\scriptsize
\centering
\begin{tabular}{lcccccc}
     \toprule
     Local & 3D-P*& & &3D-P*&3D-P*&3D-P\\ 
     Global& &2D-PI-sup&2D-PI-rs2&2D-PI-sup&2D-PI-rs2&LR\\
     \midrule
      ACC&$0.85\pm 0.06$&$ 0.76\pm 0.02$&$0.75\pm 0.01$&$0.86\pm 0.06$&$0.86\pm 0.05$&$\boldsymbol{0.88\pm 0.04}$\\
      AUC&$0.89\pm 0.05$&$ 0.78\pm 0.02$&$ 0.78\pm 0.02$&$  0.89\pm 0.04$&$  0.89\pm 0.04$&$\boldsymbol{0.93\pm 0.05}$\\
      APS&$0.92\pm 0.03$&$ 0.78\pm 0.03$&$ 0.79\pm 0.04$&$ 0.91\pm 0.03$&$ 0.91\pm 0.04$&$\boldsymbol{0.95\pm 0.03}$\\
      Recall&$0.87\pm 0.08$&$ 0.88\pm 0.08$&$ 0.86\pm 0.1$&$ \boldsymbol{0.89\pm 0.06}$&$ 0.88\pm 0.08$&$\boldsymbol{0.89\pm 0.07}$\\
      Precision&$0.87\pm0.04$&$ 0.74\pm 0.02$&$ 0.74\pm0.03$&$ 0.86\pm 0.05$&$0.87\pm 0.03$&$\boldsymbol{0.90\pm 0.02}$\\
     \bottomrule
\end{tabular}
\caption{Classification performance using local (patches) and/or global (full image) data. Abbreviations: 2D/3D: 2/3D-CNN, P: All patches, P*: Best single patch, PI-sup: Superlevel PI, PI-rs2: factor two resampled PI.}
\label{tab:results}
\end{table*}

\section{Discussion and Conclusions}
The macroscopic and microscopic brain changes related to AD provide a strong incentive for the combination of local and global imaging features for AD detection.
Here, we realised this through the combination of a 3D--CNN on a single, high-resolution image patch, in combination with either TDA or a LR ensemble model. We applied state-of-the-art pipelines for image preprocessing, topological descriptor generation, and investigated a number of preprocessing steps for the TDA part of the study. Yet, TDA did not improve classification performance in this preliminary analysis, despite encouraging results from ensemble model ii) which was competitive to previous approaches \citep{Oh2019, Pan2020, Jo2019}, motivating the merit of combining features from multiple brain locations. Our topological feature generation, despite the use of standard pipelines, may benefit from further refinements to capture more representative global connectivity features. We plan on investigating methods of filtering out topological noise, e.g. through image smoothing, or the use of different data descriptors. 
Independent of the ensemble mode used, our approach is highly parallelizable, fast ($1~s/0.1~s$ per epoch for 3D/2D--CNN training), and based on a simple architecture that permits interpretability analysis to elucidate the underlying biological hallmarks driving classification---a task we want to pursue in the future. AD--CN classification may be considered a simple task with limited clinical application given reliable other means of classification such as neuropsycological evaluation. However, we deliberately chose this imaging-based classification task to serve as a starting point for future evaluations as it is not prone to label uncertainties and can serve as input for transfer learning to classify mildly cognitively impaired patients vs AD. Finally, an important comparison will be that with a 3D--CNN trained on the full, high-resolution MRI image. Insufficient GPU memory currently limits this baseline comparison for us, which stresses the importance of providing \emph{scalable} solutions for high-resolution 3D data analysis using CNNs. We attempted training a 3D--CNN on a down-sampled ($\mathbin{{97}{\times}{115}{\times}{97}}$) image, but were restricted to a small ($<5$) batch size and slow training ($60~s$ per epoch) on the same hardware, preventing proper hyperparameter selection. Due to the large image volume, identification of localized disease markers may also require a larger ratio of training examples to model parameters than available in this study. 
A clear advantage of the presented patch-based implementation is its scalability, permitting parallelized training on different image patches that is easily conductable using state-of-the-art GPU servers with standard GPU memory. We suggest that the presented architecture may easily translate to other clinical indications employing image-based classification and, following further optimization, could help to tackle the trade-off between computational cost and imaging details retained for classification. 
 
\section*{Acknowledgements}
 This work was partially funded and erc~(S.C.B., C.R.J); Ambizione Grant \#PZ00P3186101, C.R.J). The content provided here is solely the responsibility of the authors and does not necessarily represent the official views of the funding agencies. The funders had no role in study design, data collection \& analysis, decision to publish, or preparation of the manuscript. Data collection and sharing for this project was funded by the Alzheimer's Disease Neuroimaging Initiative (ADNI) (National Institutes of Health Grant U01 AG024904) and DOD ADNI (Department of Defense award number W81XWH-12-2-0012). ADNI is funded by the National Institute on Aging, the National Institute of Biomedical Imaging and Bioengineering, and through generous contributions from the following: AbbVie, Alzheimer's Association; Alzheimer's Drug Discovery Foundation; Araclon Biotech; BioClinica, Inc.; Biogen; Bristol-Myers Squibb Company; CereSpir, Inc.; Cogstate; Eisai Inc.; Elan Pharmaceuticals, Inc.; Eli Lilly and Company; EuroImmun; F. Hoffmann-La Roche Ltd and its affiliated company Genentech, Inc.; Fujirebio; GE Healthcare; IXICO Ltd.;Janssen Alzheimer Immunotherapy Research \& Development, LLC.; Johnson \& Johnson Pharmaceutical Research \& Development LLC.; Lumosity; Lundbeck; Merck \& Co., Inc.;Meso Scale Diagnostics, LLC.; NeuroRx Research; Neurotrack Technologies; Novartis Pharmaceuticals Corporation; Pfizer Inc.; Piramal Imaging; Servier; Takeda Pharmaceutical Company; and Transition Therapeutics. The Canadian Institutes of Health Research is providing funds to support ADNI clinical sites in Canada. Private sector contributions are facilitated by the Foundation for the National Institutes of Health (www.fnih.org). The grantee organization is the Northern California Institute for Research and Education, and the study is coordinated by the Alzheimer's Therapeutic Research Institute at the University of Southern California. ADNI data are disseminated by the Laboratory for Neuro Imaging at the University of Southern California.

\bibliography{main}

\begin{thebibliography}{19}
\providecommand{\natexlab}[1]{#1}
\providecommand{\url}[1]{\texttt{#1}}
\expandafter\ifx\csname urlstyle\endcsname\relax
  \providecommand{\doi}[1]{doi: #1}\else
  \providecommand{\doi}{doi: \begingroup \urlstyle{rm}\Url}\fi

\bibitem[Adams et~al.(2017)Adams, Emerson, Kirby, Neville, Peterson, Shipman,
  Chepushtanova, Hanson, Motta, and Ziegelmeier]{Adams-PI}
Henry Adams, Tegan Emerson, Michael Kirby, Rachel Neville, Chris Peterson,
  Patrick Shipman, Sofya Chepushtanova, Eric Hanson, Francis Motta, and Lori
  Ziegelmeier.
\newblock Persistence images: a stable vector representation of persistent
  homology.
\newblock \emph{J. Mach. Learn. Res.}, 18:\penalty0 Paper No. 8, 35, 2017.
\newblock ISSN 1532-4435.

\bibitem[Ahmed et~al.(2019)Ahmed, Choi, Lee, Kim, Kwon, Lee, and
  Jung]{Ahmed2019}
Samsuddin Ahmed, Kyu~Yeong Choi, Jang~Jae Lee, Byeong~C. Kim, Goo~Rak Kwon,
  Kun~Ho Lee, and Ho~Yub Jung.
\newblock {Ensembles of Patch-Based Classifiers for Diagnosis of Alzheimer
  Diseases}.
\newblock \emph{IEEE Access}, 7:\penalty0 73373--73383, 2019.
\newblock ISSN 21693536.
\newblock \doi{10.1109/ACCESS.2019.2920011}.

\bibitem[Edelsbrunner and Harer(2010)]{Edelsbrunner-Harer-10}
Herbert Edelsbrunner and John~L. Harer.
\newblock \emph{Computational topology}.
\newblock American Mathematical Society, Providence, RI, 2010.
\newblock ISBN 978-0-8218-4925-5.
\newblock \doi{10.1090/mbk/069}.
\newblock An introduction.

\bibitem[Hofer et~al.(2017)Hofer, Kwitt, Niethammer, and Uhl]{Hofer-DLwTS}
Christoph Hofer, Roland Kwitt, Marc Niethammer, and Andreas Uhl.
\newblock Deep learning with topological signatures.
\newblock In I.~Guyon, U.~V. Luxburg, S.~Bengio, H.~Wallach, R.~Fergus,
  S.~Vishwanathan, and R.~Garnett, editors, \emph{Advances in Neural
  Information Processing Systems 30}, pages 1634--1644. Curran Associates,
  Inc., 2017.

\bibitem[Jin et~al.(2019)Jin, Xu, Zhao, Hu, Yang, Liu, Jiang, and Liu]{Jin2019}
Dan Jin, Jian Xu, Kun Zhao, Fangzhou Hu, Zhengyi Yang, Bing Liu, Tianzi Jiang,
  and Yong Liu.
\newblock {Attention-based 3D Convolutional Network for Alzheimer's Disease
  Diagnosis and Biomarkers Exploration}.
\newblock In \emph{2019 IEEE 16th International Symposium on Biomedical Imaging
  (ISBI 2019)}, pages 1047--1051. IEEE, 2019.
\newblock ISBN 978-1-5386-3641-1.
\newblock \doi{10.1109/ISBI.2019.8759455}.

\bibitem[Jo et~al.(2019)Jo, Nho, and Saykin]{Jo2019}
Taeho Jo, Kwangsik Nho, and Andrew~J. Saykin.
\newblock Deep learning in alzheimer's disease: Diagnostic classification and
  prognostic prediction using neuroimaging data.
\newblock \emph{Frontiers in Aging Neuroscience}, 11:\penalty0 220, 2019.
\newblock ISSN 1663-4365.
\newblock \doi{10.3389/fnagi.2019.00220}.

\bibitem[Korolev et~al.(2017)Korolev, Safiullin, Belyaev, and
  Dodonova]{Korolev2017}
Sergey Korolev, Amir Safiullin, Mikhail Belyaev, and Yulia Dodonova.
\newblock {Residual and plain convolutional neural networks for 3D brain MRI
  classification}.
\newblock \emph{Proceedings - International Symposium on Biomedical Imaging},
  pages 835--838, 2017.
\newblock ISSN 19458452.
\newblock \doi{10.1109/ISBI.2017.7950647}.

\bibitem[Lin et~al.(2018)Lin, Tong, Gao, Guo, Du, Yang, Guo, Xiao, Du, and
  Qu]{Lin2018}
Weiming Lin, Tong Tong, Qinquan Gao, Di~Guo, Xiaofeng Du, Yonggui Yang, Gang
  Guo, Min Xiao, Min Du, and Xiaobo Qu.
\newblock {Convolutional Neural Networks-Based MRI Image Analysis for the
  Alzheimer's Disease Prediction From Mild Cognitive Impairment}.
\newblock \emph{Frontiers in Neuroscience}, 12\penalty0 (NOV):\penalty0 1--13,
  2018.
\newblock ISSN 1662-453X.
\newblock \doi{10.3389/fnins.2018.00777}.

\bibitem[Liu et~al.(2018{\natexlab{a}})Liu, Zhang, Adeli, and Shen]{Liu2018b}
Mingxia Liu, Jun Zhang, Ehsan Adeli, and Dinggang Shen.
\newblock {Landmark-based deep multi-instance learning for brain disease
  diagnosis}.
\newblock \emph{Medical Image Analysis}, 43:\penalty0 157--168,
  2018{\natexlab{a}}.
\newblock ISSN 13618423.
\newblock \doi{10.1016/j.media.2017.10.005}.

\bibitem[Liu et~al.(2018{\natexlab{b}})Liu, Zhang, Nie, Yap, and Shen]{Liu2018}
Mingxia Liu, Jun Zhang, Dong Nie, Pew-thian Yap, and Dinggang Shen.
\newblock {Anatomical Landmark Based Deep Feature Representation for MR Images
  in Brain Disease Diagnosis}.
\newblock \emph{IEEE Journal of Biomedical and Health Informatics}, 22\penalty0
  (5):\penalty0 1476--1485, 2018{\natexlab{b}}.
\newblock ISSN 2168-2194.
\newblock \doi{10.1109/JBHI.2018.2791863}.

\bibitem[Oh et~al.(2019)Oh, Chung, Kim, Kim, and Oh]{Oh2019}
Kanghan Oh, Young~Chul Chung, Ko~Woon Kim, Woo~Sung Kim, and Il~Seok Oh.
\newblock {Classification and Visualization of Alzheimer's Disease using
  Volumetric Convolutional Neural Network and Transfer Learning}.
\newblock \emph{Scientific Reports}, 9\penalty0 (1):\penalty0 1--16, 2019.
\newblock ISSN 20452322.
\newblock \doi{10.1038/s41598-019-54548-6}.

\bibitem[Pan et~al.(2020)Pan, Zeng, Jia, Huang, Frizzell, and Song]{Pan2020}
Dan Pan, An~Zeng, Longfei Jia, Yin Huang, Tory Frizzell, and Xiaowei Song.
\newblock {Early Detection of Alzheimer's Disease Using Magnetic Resonance
  Imaging: A Novel Approach Combining Convolutional Neural Networks and
  Ensemble Learning}.
\newblock \emph{Frontiers in Neuroscience}, 14:\penalty0 259, 2020.
\newblock ISSN 1662-453X.
\newblock \doi{10.3389/fnins.2020.00259}.

\bibitem[Poulin et~al.(2011)Poulin, Dautoff, Morris, Barrett, and
  Dickerson]{Poulin2011}
St{\'{e}}phane~P. Poulin, Rebecca Dautoff, John~C. Morris, Lisa~Feldman
  Barrett, and Bradford~C. Dickerson.
\newblock {Amygdala atrophy is prominent in early Alzheimer's disease and
  relates to symptom severity}.
\newblock \emph{Psychiatry Research - Neuroimaging}, 194\penalty0 (1):\penalty0
  7--13, 2011.
\newblock ISSN 09254927.
\newblock \doi{10.1016/j.pscychresns.2011.06.014}.

\bibitem[Rieck et~al.(2020)Rieck, Yates, Bock, Borgwardt, Wolf, Turk-Browne,
  and Krishnaswamy]{Rieck20}
Bastian Rieck, Tristan Yates, Christian Bock, Karsten Borgwardt, Guy Wolf,
  Nicholas Turk-Browne, and Smita Krishnaswamy.
\newblock Uncovering the topology of time-varying {fMRI} data using cubical
  persistence.
\newblock \emph{arXiv pre-print arXiv:2006.07882}, 2020.

\bibitem[Rosenbloom and Pfefferbaum(2008)]{rosenbloom2008magnetic}
Margaret~J Rosenbloom and Adolf Pfefferbaum.
\newblock Magnetic resonance imaging of the living brain: evidence for brain
  degeneration among alcoholics and recovery with abstinence.
\newblock \emph{Alcohol Research \& Health}, 2008.

\bibitem[Scheff et~al.(2006)Scheff, Price, Schmitt, and
  Mufson]{scheff2006hippocampal}
Stephen~W Scheff, Douglas~A Price, Frederick~A Schmitt, and Elliott~J Mufson.
\newblock Hippocampal synaptic loss in early alzheimer's disease and mild
  cognitive impairment.
\newblock \emph{Neurobiology of aging}, 27\penalty0 (10):\penalty0 1372--1384,
  2006.
\newblock \doi{10.1016/j.neurobiolaging.2005.09.012}.

\bibitem[Serrano-Pozo et~al.(2011)Serrano-Pozo, Frosch, Masliah, and
  Hyman]{serrano2011neuropathological}
Alberto Serrano-Pozo, Matthew~P Frosch, Eliezer Masliah, and Bradley~T Hyman.
\newblock Neuropathological alterations in alzheimer disease.
\newblock \emph{Cold Spring Harbor perspectives in medicine}, 1\penalty0
  (1):\penalty0 a006189, 2011.
\newblock \doi{10.1101/cshperspect.a006189}.

\bibitem[Suemoto et~al.(2017)Suemoto, Ferretti-Rebustini, Rodriguez, Leite,
  Soterio, Brucki, Spera, Cippiciani, Farfel, Chiavegatto~Filho,
  et~al.]{suemoto2017neuropathological}
Claudia~K Suemoto, Renata~EL Ferretti-Rebustini, Roberta~D Rodriguez, Renata~EP
  Leite, Luciana Soterio, Sonia~MD Brucki, Raphael~R Spera, Tarcila~M
  Cippiciani, Jose~M Farfel, Alexandre Chiavegatto~Filho, et~al.
\newblock Neuropathological diagnoses and clinical correlates in older adults
  in brazil: A cross-sectional study.
\newblock \emph{PLoS medicine}, 14\penalty0 (3):\penalty0 e1002267, 2017.
\newblock \doi{10.1371/journal.pmed.1002267}.

\bibitem[Winblad et~al.(2016)Winblad, Amouyel, Andrieu, Ballard, Brayne,
  Brodaty, Cedazo-Minguez, Dubois, Edvardsson, Feldman,
  et~al.]{winblad2016defeating}
Bengt Winblad, Philippe Amouyel, Sandrine Andrieu, Clive Ballard, Carol Brayne,
  Henry Brodaty, Angel Cedazo-Minguez, Bruno Dubois, David Edvardsson, Howard
  Feldman, et~al.
\newblock Defeating alzheimer's disease and other dementias: a priority for
  european science and society.
\newblock \emph{The Lancet Neurology}, 15\penalty0 (5):\penalty0 455--532,
  2016.
\newblock \doi{10.1016/S1474-4422(16)00062-4}.

\end{thebibliography}

\clearpage

\appendix

\section{Supplements}

\subsection{Preprocessing of MRI data}
\label{apd:preprocessing}
We included all T1-weighted MRI images from ADNI1, 2, 3, and GO, which were captured and preprocessed by ADNI.
Results included in our work come from preprocessing performed using \texttt{fMRIPrep} 20.1.1, a Nipype 1.5.0 based tool.
All MRIs were corrected for intensity non-uniformity (INU) with \texttt{N4BiasFieldCorrection}, distributed with ANTs 2.2.0, and used as T1w-reference throughout the workflow.
The T1w-reference was then skull-stripped with a \texttt{Nipype} implementation of
the \texttt{antsBrainExtraction.sh} workflow (from ANTs), using \texttt{OASIS30ANTs} as a target template.
Volume-based spatial normalisation to a standard coordinate space~(\texttt{MNI152NLin2009cAsym}) was performed through
nonlinear registration with \texttt{antsRegistration}, using brain-extracted versions of both T1w reference and the T1w template.
We slected the template `ICMB 152 Nonlinear Asymmetrical Template Version 2009c' for spatial normalisation.

Many internal operations of \texttt{fMRIPrep} use the
\texttt{Nilearn} library, version 0.6.2, 
mostly within the functional processing workflow.
For more details of the pipeline, please refer to \href{https://fMRIPrep.readthedocs.io/en/latest/workflows.html}{the official documentation of \texttt{fMRIPrep}}. Preprocessing was finalised by intensity normalisation of the extracted and MNI space registered brain images. 

\end{document}